\documentclass[10pt,twocolumn,letterpaper]{article}

\usepackage{iccv}
\usepackage{times}
\usepackage{epsfig}
\usepackage{graphicx}
\usepackage{amsmath}
\usepackage{amssymb}
\usepackage{xcolor}
\usepackage{multirow}
\usepackage{booktabs}
\usepackage{capt-of}

\usepackage[title]{appendix}

\newcommand{\CLS}[0]{\mathit{CLS}}
\usepackage[breaklinks=true,bookmarks=false]{hyperref}

\iccvfinalcopy % *** Uncomment this line for the final submission

\ificcvfinal\fi

\usepackage[export]{adjustbox}
\usepackage{fontawesome}
\usepackage{tikz}
\usetikzlibrary{fit}
\begin{document}

%%%%%%%%% TITLE
\title{Unsupervised Open-Vocabulary Object Localization in Videos}

\author{Ke Fan\textsuperscript{1,$\ast$}, 
Zechen Bai\textsuperscript{2,$\ast$},
Tianjun Xiao\textsuperscript{2},
Dominik Zietlow\textsuperscript{2},
Max Horn\textsuperscript{2},
Zixu Zhao\textsuperscript{2},\\
Carl-Johann Simon-Gabriel\textsuperscript{2},
Mike Zheng Shou\textsuperscript{3},
Francesco Locatello\textsuperscript{2},\\
Bernt Schiele\textsuperscript{2},
Thomas Brox\textsuperscript{2},
Zheng Zhang\textsuperscript{2,$\dagger$},
Yanwei Fu\textsuperscript{1,$\dagger$},
Tong He\textsuperscript{2}
\\
\textsuperscript{1}{Fudan University} \quad
\textsuperscript{2}{Amazon Web Services} \quad 
\textsuperscript{3}{National University of Singapore}\\
{\tt\small kfan21@m.fudan.edu.cn, \{baizeche, tianjux, zhaozixu, zhaz, htong\}@amazon.com}\\
{\tt\small\{zietld, hornmax, cjsg, locatelf, bschiel, brox\}@amazon.de}\\
{\tt\small mikeshou@nus.edu.sg, yanweifu@fudan.edu.cn}\\
}

\twocolumn[{%
\renewcommand\twocolumn[1][]{#1}%
\maketitle
\vspace{-25pt}
\begin{center}
    \centering
    \includegraphics[width=0.86\textwidth]{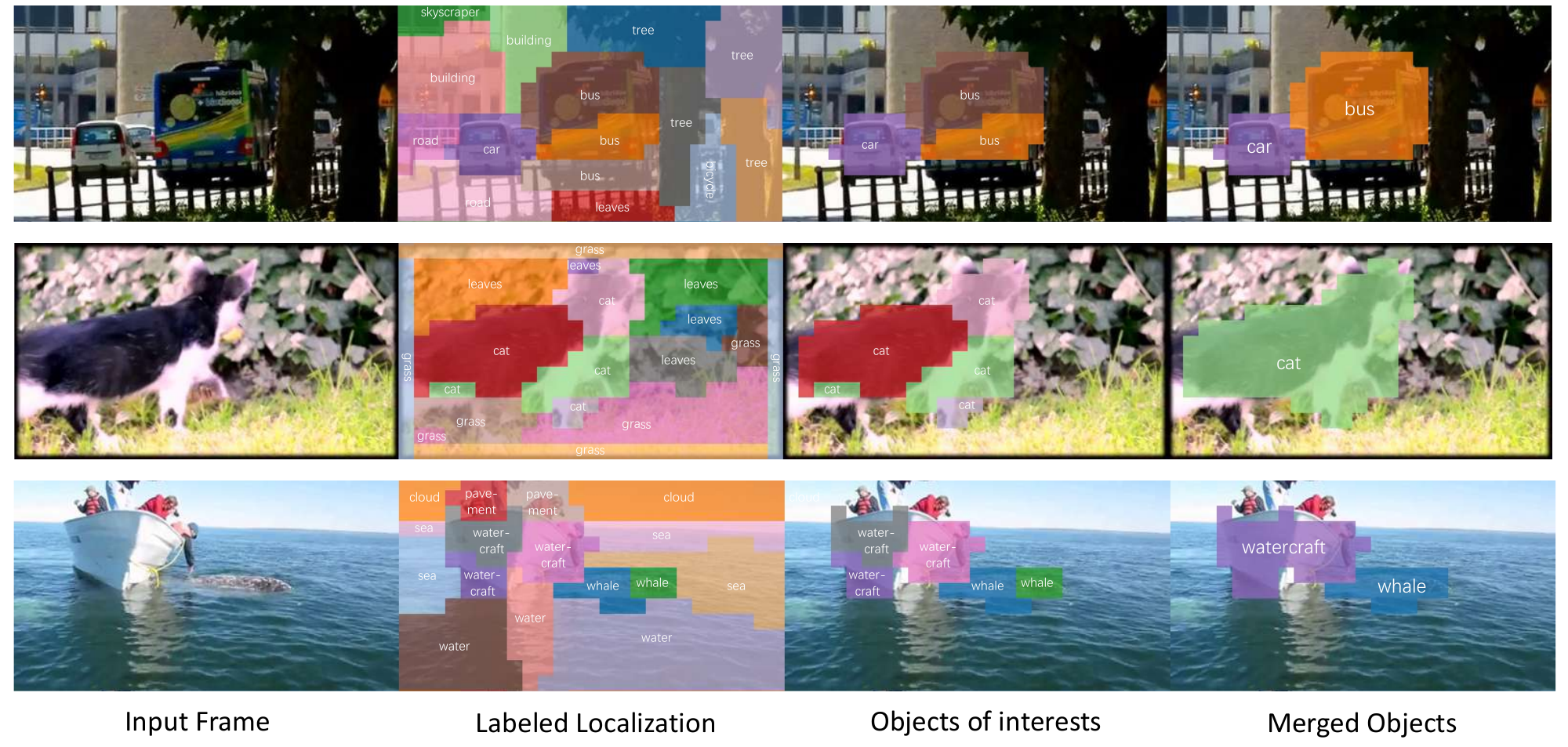}
    \captionof{figure}{We propose an unsupervised approach to localize and name objects in real-world videos. The approach uses slot attention in feature space to localize tubes (second column), assigns text to the slot features via a CLIP model that was modified to allow local feature alignment (third column), and finally merges slots that overlap in text space (last column).}\label{fig:teaser}
\end{center}
}]

\makeatletter
\def\blfootnote{\gdef\@thefnmark{}\@footnotetext}
\makeatother

\blfootnote{$\ast$ Equal contribution. Ke Fan is the first intern author, Zechen Bai is the first FTE author, they contributed equally. Work down during Ke Fan's internship in AWS Shanghai AI Lab}
\blfootnote{$\dagger$ Corresponding authors. Yanwei Fu and Ke Fan are with the School of Data Science, Fudan University.}

\ificcvfinal\thispagestyle{empty}\fi

%%%%%%%%% BODY TEXT
\begin{abstract}
    In this paper, we show that recent advances in video representation learning and pre-trained vision-language models allow for substantial improvements in self-supervised video object localization. We propose a method that first localizes objects in videos via an object-centric approach with slot attention and then assigns text to the obtained slots. The latter is achieved by an unsupervised way to read localized semantic information from the pre-trained CLIP model. The resulting video object localization is entirely unsupervised apart from the implicit annotation contained in CLIP, and it is effectively the first unsupervised approach that yields good results on regular video benchmarks. 
    Project Page:\url{https://github.com/amazon-science/object-centric-vol}
\end{abstract}
\section{Introduction}

Deep learning has demonstrated remarkable performance in object-detection and -recognition, both on images and videos. Most models are trained via supervised learning, which requires expensive manual annotation of training data. Such approaches also run the risk of overfitting to specific distributional characteristics of the training data, and may not generalize well beyond that. Self-supervised learning methods are a promising approach to overcome this problem. They can learn from vast amounts of unlabeled visual data and unlock the full potential of deep learning, for example in video understanding.

Recently, a lot of progress was made in training segmentation models without supervision. 
Caron et al.~\cite{Caron2021DINO} demonstrated that vision transformers, combined with self-supervised learning, can create a feature space in which patterns of the same object class cluster. This feature space was used in a series of papers to learn unsupervised segmentation models~\cite{Melaskyriazi2022DeepSpectral,Zadaianchuk2022COMUS,dinosaur}. 
In another line of research, Locatello et al.~\cite{Locatello2020SlotAttention} proposed an approach, where so-called \emph{slots} localize individual objects in the scene and represent their visual patterns and properties. This type of approach belongs to the field of \emph{object-centric learning (OCL)}. So far, much of the OCL literature has focused on static images. Only recently it became possible to obtain good results not only on synthetic but also on real-world datasets~\cite{dinosaur}. Few papers have tried to apply a slot-based OCL approach to video data \cite{kipf2022conditional, elsayed2022savi++}. These works scale to real-world data by adding some form of weak annotation.

In contrast, this paper proposes an OCL pipeline for \emph{real-world} video data with two main goals: (1) partitioning the videos into spatially, temporally, and semantically meaningful groups \emph{without using any labeled training data}; and (2) labeling those groups using an off-the-shelf vision-language model, such as CLIP~\cite{Radford2021clip}. 
Since CLIP was trained to align text with global image features, it is initially unable to align text with local features from individual objects. However, CLIP can be fine-tuned on image-text data to make it align the local patterns to text~\cite{Mukhoti2022pacl}.
In this paper, we show that paired image-text data is not necessary to allow for such alignment. We propose a self-supervised objective to fine-tune the last layer of the CLIP vision transformer, which trains on image data only. 

\begin{figure*}[htb]
    \centering
    \includegraphics[width=0.95\linewidth]{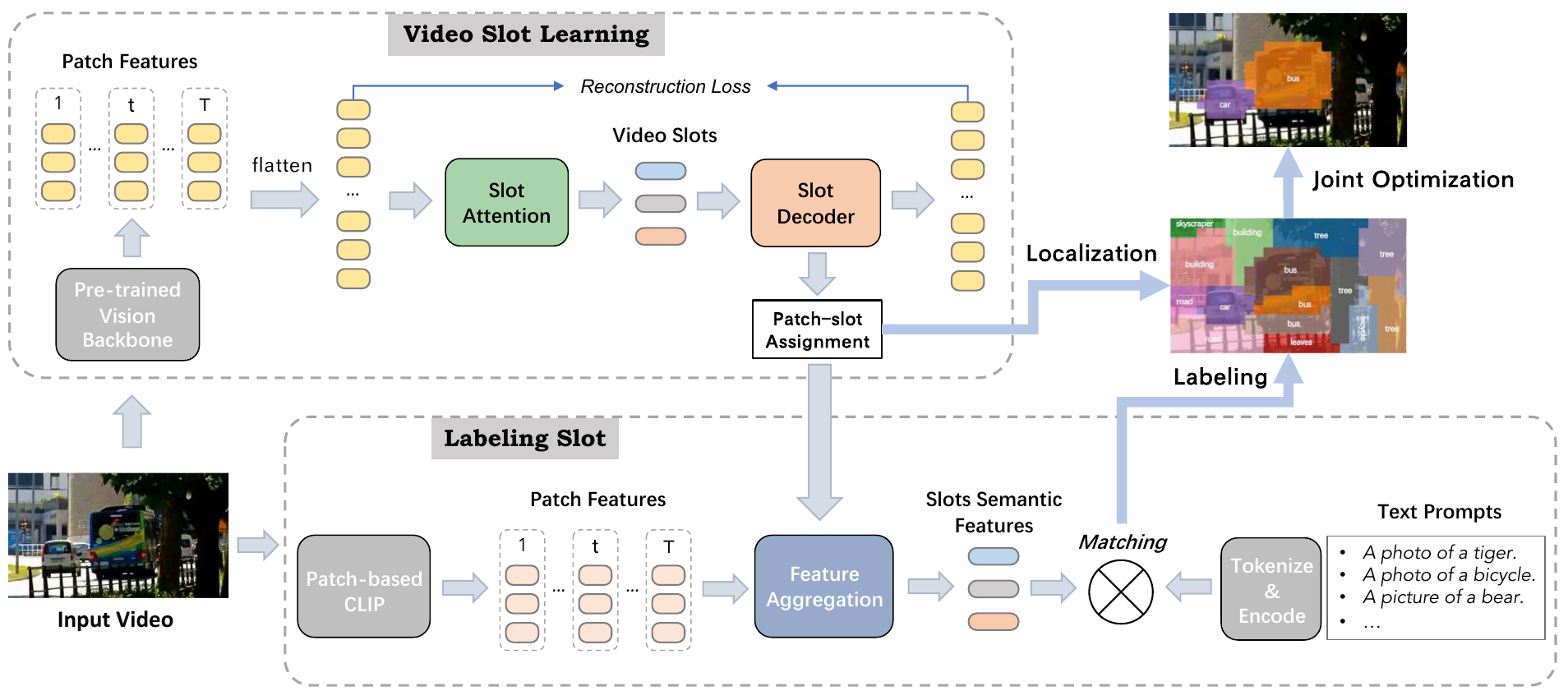}
    \caption{\textbf{Proposed framework.} Given an input video, we first localize objects by slot attention with a video encoder pretrained with self-supervision. Next, we extract semantic features for each slot by a patch-based CLIP finetuned from its vanilla version. Then, slots are named by matching slot semantic features to text features from a curated list of text prompts. Finally, the named slots are optimized to alleviate over-segmentation caused by part-whole hierarchies.}
    \label{fig:framework}
    \vspace{-1em}
\end{figure*}

\noindent\textbf{Overview.}
The overall framework is composed of three parts, as depicted in Figure~\ref{fig:framework}. The individual processing stages are visualized in Figure~\ref{fig:teaser}. The first part yields bottom-up object-centric video representations (slots) that localize the regions in the video with a spatio-temporal grouping model (Section~\ref{sec:oclocalization}). The second part adapts the off-the-shelf image-text alignment model CLIP~\cite{Radford2021clip} to assign a text to the video slots (Section~\ref{sec:vslot_naming}). Finally, a merging procedure leverages overlapping information from the text and the image to improve both localization and labeling (Section~\ref{sec:postprocess}). Our contributions are twofold:
\begin{itemize}
    \item We provide the first approach that localizes objects with spatio-temporal consistency in real-world videos, without labelled training data.%any manual labeling. 
    \item We assign text labels to video slots using a pre-trained CLIP model without additional supervised fine-tuning. 
\end{itemize}
\section{Related Work}

\noindent\textbf{Object-centric localization.}
Our research contributes to the field of object-centric learning,  which involves extracting individual objects from visual input using neural networks with inductive biases.  Some approaches, such as using an information bottleneck with pixel-level reconstruction~\cite{Singh2022STEVE,Locatello2020SlotAttention, xie2021unsupervised} or examining perceptual similarity through self-supervised representation pretraining~\cite{Engelcke2020GENESIS,dittadi2021generalization}, have been previously explored.
Our contribution is introducing temporal coherence into the mix, which improves object emergence and enhances the objectness signal by leveraging the Common Fate principle. Furthermore, there have been recent advances in object-centric learning for video applications~\cite{kipf2022conditional, elsayed2022savi++,Singh2022STEVE,dom}. These approaches improve upon previous image-level methods by incorporating temporal modules into the network architecture. However, they either require additional annotation signals in the form of the first frame and optical flow, or they rely on complex modules to enable temporal tracking or matching of object slots in neighboring frames. In comparison, our method does not require any additional signals or specifically designed network modules, yet still achieves competitive performance on real-world videos.

\noindent\textbf{CLIP for low-level vision tasks.}
Recent advances in vision-language models have been driven by large-scale  Contrastive Language-Image Pre-Training (CLIP) model~\cite{Radford2021clip}. While CLIP has shown promising results in low-level vision tasks, its training setup has been criticized for lacking localization ability~\cite{ Mukhoti2022pacl, Liang2022ovsemantic,Du2022prompt_open_vocab,Xu2022GroupViT}. Some solutions involve using gradient activation maps~\cite{Selvaraju2017gradcam} or extra annotation to finetune CLIP~\cite{Xu2022GroupViT}. Our contribution is decoupling the tasks of object-centric learning and vision-language, allowing us to finetune CLIP with only a moderate \textit{unlabeled} dataset for semantic feature extraction, without the need for additional annotations. This approach stands in contrast to previous works that require additional training annotations for both localization and semantic labeling using a single vision-language module~\cite{Xu2022GroupViT}.

\section{Video slot learning}\label{sec:oclocalization}

\noindent\textbf{Problem setup.}
Given a video $V\in\mathbb{R}^{T\times H\times W\times 3}$, the task is to extract $K$ video slots $S_{\text{video}}\in\mathbb{R}^{K\times D_\text{slot}}$ that bind to object features in the video, as well as a video alpha mask $\alpha\in\mathbb{R}^{K\times T\times H\times W}$ used to segment the video into $K$ tubes.

\noindent\textbf{Self-supervised video encoder.}
For a pre-defined $p\times p\times 1$ patch size, we tokenize the video $V$ and flatten the spatio-temporal dimension to a sequence of tokens $\text{tn}_V=\text{Tokenize}(V) \in\mathbb{R}^{T\times H' \times W' \times D}$, where $H'= \lceil\frac{H}{p}\rceil$ and $W' = \lceil\frac{W}{p}\rceil$.
After adding spatio-temporal positional encoding, the tokens are processed by a video encoder $\text{E}_\text{V}$ as $D$-dimensional video token features, $F=\text{E}_\text{V}(V)\in \mathbb{R}^{T \times H' \times W' \times D}$. 
The video encoder consists of a stack of transformer blocks, an architecture that has been shown effective for self-supervised vision representation learning without annotation~\cite{Caron2021DINO,tong2022videomae,Feichtenhofer2022maskedmae}. 

\noindent\textbf{Learning video slots.}
We generalize DINOSAUR~\cite{dinosaur}, a state-of-the-art object-centric learning method for \emph{static} images, to videos.
DINOSAUR passes the tokens of an image, \textit{e.g.}, a frame at time step $t$ through a pre-trained image encoder $\text{E}_\text{I}$ and obtains $D$-dimensional image features $f_t=\text{E}_\text{I}(\text{tn}_{V_t})\in\mathbb{R}^{H^\prime\times W^\prime\times D}$.
It then flattens the spatial dimensions of $f_t$ and applies slot-attention~\cite{Locatello2020SlotAttention} as follows:
$$
S_{\text{img},t} = \operatorname{SlotAttention}(f_{t}.\text{reshape}(H^\prime W^\prime, D)).
$$
$S_{\text{img},t}$ effectively represents image slots, each corresponding to a group of pixels that are supposed to cover an object.

To use DINOSAUR on videos, a straightforward approach would be to run it separately on each frame of the video.
However, this require to run slot-attention $T$ times and results in poor temporal consistency among slots in multiple frames. 
Instead, we run slot-attention only once on the entire spatio-temporal features ${F \in \mathbb{R}^{T \times H' \times W' \times D}}$ of the video.
Specifically, we flatten the spatial and temporal dimensions and extract video slots for all frames as follows:
$$
S_{\text{video}} = \operatorname{SlotAttention}(F.\text{reshape}(T H^\prime W^\prime, D))\in \mathbb{R}^{K\times D_\text{slot}}.
$$
$S_{\text{video}}$ are video slots that capture video features.
Since patches from a same object tend to have similar features across subsequent frames, the slot attention mechanism tends to group them into a same slot, which ensures temporal consistency \emph{by design}.

\noindent\textbf{Slot decoder and training loss.}
With $S_{\text{video}}$, each slot $S_i\in\mathbb{R}^{D_\text{slot}}$ is first boardcasted to a 3D volume $V_i\in\mathbb{R}^{T\times H\times W\times D_\text{slot}}$ and added to a spatio-temporal positional encoding $pe$, then a 2-layer Multi-Layer Perceptron (MLP) is applied on each position individually:
$$
y_i, \alpha_i = \operatorname{MLP}(V_i + pe)
$$
where we have $\mathbf{y}=[y_1, \dots, y_K]$ with $y_i\in \mathbb{R}^{T\times H'\times W'\times D}$ as the feature reconstruction, and $\mathbf{\alpha}=[\alpha_1, \dots, \alpha_K]$ with $\alpha_i\in\mathbb{R}^{T\times H'\times W'}$ as the alpha mask for the patch-to-slot assignment weight.
Specifically, let $\alpha_{\cdot,j} = [\alpha_{1,j}, \dots, \alpha_{K,j}]$ be the alpha mask weight for the $j$-th patch, then $\sum_{i=1}^K \alpha_{i, j} = 1$.
We say that the $j$-th patch belongs to the $i^\ast$-th slot if $i^\ast=\arg\max_i \alpha_{i, j}$. 
Similar to previous work~\cite{Singh2022STEVE, dinosaur}, during training, we use the following reconstruction loss of video features $F$ based on the slot-attention output $\mathbf{y}$ and $\mathbf{\alpha}$:
$$L = \text{MSE}(\mathbf{y} \cdot \mathbf{\alpha}, F).$$

\section{Labeling video slots}\label{sec:vslot_naming}
\noindent\textbf{Problem Setup.}
Given a video $V\in\mathbb{R}^{T\times H\times W\times 3}$, its alpha mask $\alpha\in\mathbb{R}^{ K\times T\times H'\times W'}$ and a list of $N$ potential labels and corresponding features $f_\text{text}\in\mathbb{R}^{N\times D_\text{sem}}$, the task is to come up with a slot-to-label assignment matrix $A\in\mathbb{R}^{K\times N}$

\noindent\textbf{A naive baseline with CLIP.}
CLIP~\cite{Radford2021clip} is a vision-language model pre-trained on an internet-scale dataset.
For each image-text pair, CLIP summarizes the text and vision features in tokens $\CLS_t$ and $\CLS_v$ and aligns them using contrastive learning.
The tokens $\CLS$ are constructed using an N-stack transformer.
Since CLIP can be used to label almost any full image, a straightforward approach for slot labeling would be to wrap the video slots into bounding boxes, crop the regions and send each of them to CLIP.
However, this method suffers from two problems. First, this requires multiple runs of the CLIP visual encoder for one input image. Second, when a slot is non-convex in shape, its corresponding bounding box would inevitably include content that does not belong to the slot and eventually bias the features from CLIP.
In the following, we describe a more efficient and unbiased approach to slot labeling which requires minor modifications on the pre-trained CLIP model but no additional labels.

\noindent\textbf{Efficient slot labeling with semantic features.}
\looseness-1Assume there is an image semantic encoder $E_\text{sem}$ that operates on the same patch size $p$ as in Section~\ref{sec:oclocalization}.
We encode the $t$-th frame $I_t$ into a set of semantic patch features $\{P_t\}=E_\text{sem}(I_t)\in\mathbb{R}^{H'\times W'\times D_\text{sem}}$.
For each video slot, we average the patch features in this slot as the slot semantic feature $\{f_\text{slot}\}\in\mathbb{R}^{K\times D_\text{sem}}$.
Given a list of potential semantics and corresponding features $\{f_\text{text}\}\in\mathbb{R}^{N\times D_\text{sem}}$, we compute the cosine similarity between $f_\text{slot}$ and $f_\text{text}$.
Let $A$ be the matrix for slot-text similarity: $A_{ij}=\cos(f_{\text{slot},i}, f_{\text{text},j})$.
For each slot, we find the index of the text with maximum feature cosine similarity $j^\ast = \arg\max_j A_{ij}$ and label the slot accordingly.

\noindent\textbf{Local semantics from CLIP.}
Although the output of CLIP's visual encoder contains both the $\CLS_v$ token as well as image patch tokens, the contrastive loss used during pre-training only enforces alignment between the text token $\CLS_t$ with $\CLS_v$, not with the other visual tokens.
And in practice, those other patch tokens do not align with the text features $\CLS_t$, as observed by~\cite{Mukhoti2022pacl, Liang2022ovsemantic}, and confirmed by our own initial experiments.
Therefore, the visual encoder from a vanilla CLIP cannot be used without further modification as the semantic encoder $E_\text{sem}$.
To fix this issue, \cite{Mukhoti2022pacl, Liang2022ovsemantic} adapt the CLIP visual encoder to local semantic features by fine-tuning it on additional vision data with image-text pairing, for example segmentation masks \cite{Liang2022ovsemantic}.

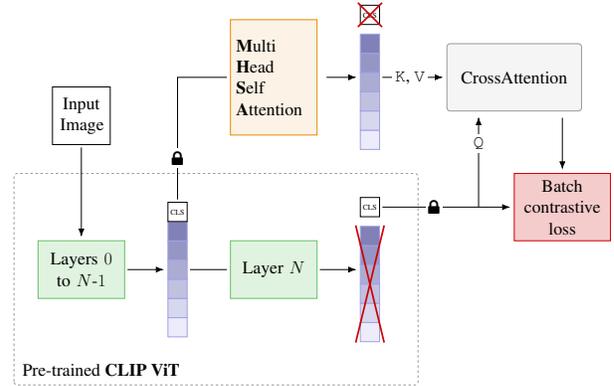
\begin{figure}[htb]
    \centering
    \adjustbox{width=\linewidth,trim=6em 0 0em 0}{
    \colorlet{mywhite}{white}
\colorlet{mygray}{white!60!black}
\colorlet{myblue}{blue!30!white!90!black}
\colorlet{myorange}{red!40!yellow!90!black}
\colorlet{mygreen}{green!60!black!60!white}
\colorlet{myred}{red!80!black}
\colorlet{mypurple}{myred!40!myblue}

\usetikzlibrary{patterns}
\usetikzlibrary{arrows.meta}
\usetikzlibrary{calc}
\usepgflibrary{shapes.misc} % LaTeX and plain TeX and pure pgf
\usepgflibrary[shapes.misc] % ConTeXt and pure pgf
\usetikzlibrary{shapes.misc} % LaTeX and plain TeX when using TikZ
\usetikzlibrary[shapes.misc] % ConTeXt when using TikZ

\tikzstyle{node}=[inner sep=0,outer sep=0]
\tikzstyle{myline}=[thick,black,shorten >=1,line width=0.4mm]
\tikzstyle{connect}=[->,myline,-{Latex[length=3.5mm]}]
\tikzstyle{every node}=[font=\huge]
    \colorlet{myblues}{red!10!myblue}
\tikzset{
  node 1/.style={node,black,draw=black,fill=mywhite!25},
  node 2/.style={node,black,draw=mygray,fill=mygray!10},
  node 3/.style={node,black,draw=myblue,fill=myblue!20},
  node 4/.style={node,black,draw=mygreen,fill=mygreen!20},
  node 5/.style={node,black,draw=mygreen,fill=mygreen!20},
  node 6/.style={node,black,draw=myorange,fill=myorange!10},
  node 7/.style={node,black,draw=myred,fill=myred!20},
}

\def\featureWidth{0.8}

\def\modelWidth{0.9}
\def\modelHeight{0.6}

\def\modelHeightMSHA{1.2}

\def\operationWidth{1.4}
\def\operationHeight{0.7}

\def\dataWidth{0.6}
\def\dataHeight{0.6}

\def\lossWidth{1}
\def\lossHeight{0.7}

\def\trOutputSize{0.2}

\begin{tikzpicture}[x=4cm,y=4cm,font={\sffamily}]
\tikzset{font=\Huge}

  \newcommand{\model}[4]{
    \draw [#3, text width=100, align=center] (#1-\modelWidth/2,#2-\modelHeight/2) rectangle (#1+\modelWidth/2, #2+\modelHeight/2) node[pos=.5] {#4};
  }

  \newcommand{\modelMHSA}[4]{
    \draw [#3, text width=100, align=left] (#1-\modelWidth/2,#2-\modelHeightMSHA/2) rectangle (#1+\modelWidth/2, #2+\modelHeightMSHA/2);
    \node[text width=100, align=left, ] at (#1+0.05,#2) {#4};
  }

  \newcommand{\operation}[3]{
    \draw [rounded corners, align=center,node 2] (#1-\operationWidth/2,#2-\operationHeight/2) rectangle (#1+\operationWidth/2, #2+\operationHeight/2) node[pos=.5] {#3};
  }

  \newcommand{\transformerOutput}[6]{
    \colorlet{myColorA}{#4}
    \colorlet{myColorB}{#5}
    \foreach \y in {0,...,5} {
        \draw [node 3, fill=myColorA!\fpeval{100*(1-((\y)) / 5)}!myColorB]
        (#1 - \trOutputSize/2, #2 - \trOutputSize*3.5 + \fpeval{\y} * \trOutputSize - #6/2) rectangle 
        (#1 + \trOutputSize/2, #2 - \trOutputSize*3.5 + \fpeval{\y+1} * \trOutputSize - #6/2) node[pos=.5] {#3};
    }
        \draw [node 1]
        (#1 - \trOutputSize/2, #2 - \trOutputSize*3.5 + \fpeval{6} * \trOutputSize + #6/2) rectangle 
        (#1 + \trOutputSize/2, #2 - \trOutputSize*3.5 + \fpeval{7} * \trOutputSize + #6/2) node[pos=.5] {\small CLS};
  }

  \newcommand{\data}[3]{
    \draw [text width=50, align=center] (#1-\dataWidth/2,#2-\dataHeight/2) rectangle (#1+\dataWidth/2, #2+\dataHeight/2) node[pos=.5] {#3};
  }

  \newcommand{\patchedData}[5]{
    \colorlet{myColorA}{#4}
    \colorlet{myColorB}{#5}
    \foreach \x in {1,...,3} {
        \foreach \y in {1,...,2} {
            \draw [node 3, fill=myColorA!\fpeval{100*(1-((\x-1)*2+(\y-1)) / 5)}!myColorB]
            (#1 - \dataWidth/2 - \dataWidth/3/2 + \fpeval{\x-0.5} * \dataWidth/3, #2 - \dataHeight/2 -\dataHeight/2/2 + \fpeval{\y-0.5} * \dataHeight/2) rectangle 
            (#1 - \dataWidth/2 + \dataWidth/3/2 + \fpeval{\x-0.5} * \dataWidth/3, #2 - \dataHeight/2 +\dataHeight/2/2 + \fpeval{\y-0.5}\y * \dataHeight/2) node[pos=.5] {#3};
        }
    }
  }

  \newcommand{\loss}[3]{
    \draw [node 7,align=center] (#1-\lossWidth/2,#2-\lossHeight/2) rectangle (#1+\lossWidth/2, #2+\lossHeight/2) node[pos=.5, outer sep=0] {#3};
  }

\begin{scope}[local bounding box=transformer decoder box]
\draw[rounded corners, fill=white, draw=mygray, dashed] (1.3, -0.2) rectangle (5.5, 2) {};
\end{scope}
\node [fit=(transformer decoder box), outer sep=0pt, inner sep=0pt] (transformer decoder) {};
\node[anchor=south west,inner sep=10pt] at (transformer decoder.south west) {Pre-trained \textbf{CLIP ViT}};

\begin{scope}[local bounding box=image box]
\data{2}{2.6}{Input\\Image}
\end{scope}
\node [fit=(image box), outer sep=0pt] (image) {};

\colorlet{gradientA}{myblue!20!white}
\colorlet{gradientB}{myblue!80!black}

\begin{scope}[local bounding box=clip n1 box]
\model{2}{1}{node 5}{Layers $0$ to $N$-$1$}
\end{scope}
\node [fit=(clip n1 box), outer sep=0pt] (clip n1) {};

\begin{scope}[local bounding box=clip n1 output box]
\transformerOutput{3}{1}{}{gradientA}{gradientB}{0}
\end{scope}
\node [fit=(clip n1 output box), outer sep=0pt] (clip n1 output) {};

\begin{scope}[local bounding box=clip n box]
\model{4}{1}{node 5}{Layer $N$}
\end{scope}
\node [fit=(clip n box), outer sep=0pt] (clip n) {};

\begin{scope}[local bounding box=clip n output box]
\transformerOutput{5}{1}{}{gradientA}{gradientB}{0.1}
\end{scope}
\node [fit=(clip n output box), outer sep=0pt] (clip n output) {};
\node [cross out,draw=myred, text width=25, text height=130, line width=2] at (5, 0.85) {};

\begin{scope}[local bounding box=mhsa box]
\modelMHSA{4}{3}{node 6}{\textbf{M}ulti\\\textbf{H}ead\\\textbf{S}elf\\\textbf{A}ttention}
\end{scope}
\node [fit=(mhsa box), outer sep=0pt] (mhsa) {};

\begin{scope}[local bounding box=mhsa output box]
\transformerOutput{5}{3}{}{gradientA}{gradientB}{0.1}
\end{scope}
\node [fit=(mhsa output box), outer sep=0pt] (mhsa output) {};
\node [cross out,draw=myred, text width=20, text height=20, line width=2] at (5, 3.65) {};
\begin{scope}[local bounding box=crossattention box]
\operation{6.5}{3}{CrossAttention}
\end{scope}
\node [fit=(crossattention box), outer sep=0pt] (crossattention) {};

\begin{scope}[local bounding box=loss box]
\loss{7}{1.65}{Batch\\contrastive\\ loss}
\end{scope}
\node [fit=(loss box), outer sep=0pt] (loss) {};

\draw[connect] (image.south) -- (clip n1.north);
\draw[connect] (clip n1.east) -- (clip n1 output.west);
\draw[connect, -] (clip n1 output.east) -- (clip n.west);
\draw[connect] (clip n.east) -- (clip n output.west);

\draw[connect, -] (clip n1 output.north) |- (mhsa.west) node[pos=0.17, fill=white] () {\faLock};
\draw[connect] (mhsa.east) -- (mhsa output.west);

\draw[connect] ($(clip n output.east) +(1.0, 0.65)$) -- ($(clip n output.east)+(1.0,1.6)$) node[pos=0.7, fill=white] () {\texttt{Q}};

\draw[connect] (mhsa output.east) -- (crossattention.west) node[pos=0.45, fill=white] () {\texttt{K}, \texttt{V}};;
\draw[connect] ($(loss.north)+(0,0.55)$)  -- (loss.north);

\draw[connect] ($(clip n output.east)+(0, 0.65)$) -- (loss.west) node[pos=0.4, fill=white] () {\faLock};
\end{tikzpicture}
    }
    \caption{\textbf{Patch-based CLIP finetuning}.
    We replace the last ViT layer by a multi-head self-attention module to re-project semantic information to the new patch tokens.
    We then use cross-attention to encourage those patches that contain the main context to be similar to the $\CLS_v$ token.
    Maybe surprisingly, this suffices to get semantically meaningful patch features.
    Importantly, we do not use any labeled data during this fine-tuning step.
    }%\tong{this is a quick place holder. Aesthetic, notation and layout will be changed.}}
    \label{fig:clip_tune}
    \vspace{-2em}
\end{figure}

\noindent\textbf{Patch-based CLIP.}
\label{sec:patch_based_clip}
In contrast, we adapt the pre-trained CLIP visual encoder to a \textit{patch-based CLIP} visual encoder using only \emph{unlabeled images}, as illustrated in Figure~\ref{fig:clip_tune}.
The idea is based on this observation: the semantic-informative token $CLS_v$ is the output from a ViT's last layer, thus we can train a new function that re-projects the information from the last layer's input to new patch features.

Formally, given an input image $I$ and the pre-trained CLIP visual encoder $\text{E}_\text{CLIP}^{v}$, we consider both its output token $\CLS_v \in \mathbb{R}^{D_\text{sem}}$ as well as the patch tokens $P' \in \mathbb{R}^{M\times D_\text{sem}}$ of the ViT that were used in the penultimate layer to construct $\CLS_v$.
We replace the last layer by a new multi-head self-attention module $\text{E}_\text{MHSA}^{v}$ and obtain new patch tokens $\hat{P}=\text{E}_\text{MHSA}^{v}(P')$. 

This new module is trained with self-supervision, under the assumption that a pre-trained CLIP has its $\CLS_v$ close to the corresponding text token $\CLS_t$. We use $\CLS_v$ as a proxy to the semantic representation space. We first compute a cross-attention with queries being $\CLS_v$, while keys and values are both $\hat{P}$:

\begin{equation}
    \hat{P}_v= \text{Softmax}(\hat{P} \cdot \CLS_v^T) \cdot \hat{P}.
\end{equation}

\noindent We then minimize the InfoNCE loss~\cite{oord2018representation} between the aggregated feature $\hat{P}_v$ and $\CLS_v$ across data samples. 
\begin{equation}
    \phi_{ij}=\frac{\hat{P}_{v,i}}{|\hat{P}_{v,i}|}\cdot \frac{\CLS_j}{|\CLS_j|},
\end{equation}
\begin{equation}
    \mathcal{L}_{\text{InfoNCE}}=\frac{1}{k}\sum_{i}^{k}\log(\frac{e^{\phi_{ii}}}{\sum_{j}^{k}e^{\phi_{ij}}}).
\end{equation}
We use this scheme to adapt pre-trained CLIP patch features to the image-text joint space while training only on unlabeled image data.
We do not use any visual annotation or image-text pairs.
Intuitively, our modification of CLIP simply allows us to read out the information that was already contained in the CLIP model, but was not readily available. 

\section{Post-processing and joint optimization}\label{sec:postprocess}
The localization and text assignment procedures above already provide a good basis, but they leave room for improvements. In the following, we describe two post-processing steps that improve clustering on the text and the video data, respectively. 

\noindent\textbf{Optimizing labels with curated vocabulary.}
The similarity-based algorithm described from Section~\ref{sec:vslot_naming} requires a vocabulary of possible labels.
In principle, the patch-based CLIP is capable to embed and recognize any word that appears in its large-scale training set.
However, a too large vocabulary can lead to sub-optimal results due to synonyms and to less robust ranking performance against a too-long list of text features.

\looseness-1In practice, a good vocabulary should include both \textit{target labels} and \textit{common background labels}. Target labels are the dataset specific foreground object categories of interest. Common background labels are concepts that usually appear in the background and do not overlap with target labels.

Without prior knowledge, it is still possible that content of a slot may not correspond to any category, neither in the target, nor in common background labels.
As a result, such a slot would have features that are not similar to any text features.
We therefore leave slot $i$ ``unnamed'' if $\cos(f_{\text{slot},i}, f_{\text{text},j}) < \lambda$ for all $j$, for some constant threshold $\lambda$. In practice, we find $\lambda$ can be simply set to 0.

\noindent\textbf{Improving part-whole with name-assisted localization.}
Using the slot labels (i.e., names), we can identify slots that cover only part of an object and merge them into a new slot with the same name.

\begin{figure}[htb]
    \centering
    \includegraphics[width=0.9\linewidth]{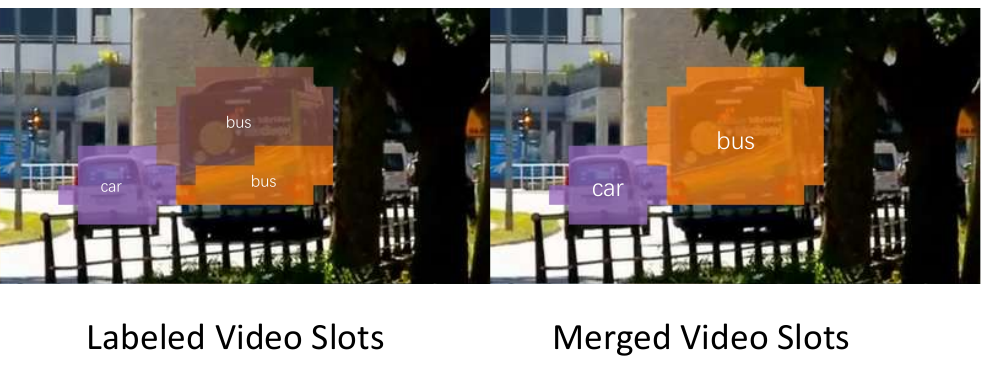}
    \caption{\textbf{Joint optimization}. The image on the left shows one frame of video slots with target names, the image on the right shows the result from merging. The two slots for the bus are merged thus better localizing the object, while they don't further merge with the slot for car since they share different semantics.}
    \label{fig:joint_opt}
\end{figure}

Specifically, for each pair of slots $S_i$ and $S_j$, we consider them as being part of the same object if they have the same label and are spatial neighbors in at least one frame.
This merging process is repeated until there's no pair of slots to merge.
Figure~\ref{fig:joint_opt} illustrates this process.
Although seemingly simple, this step is crucial to harness the part-whole issue.
It builds both parts of our pipeline so far: the localization and the semantic label information.
This concludes the description of our approach.
The next section describes our evaluation pipeline.

\begin{figure}[ht]
    \centering
    \includegraphics[width=0.99\linewidth]{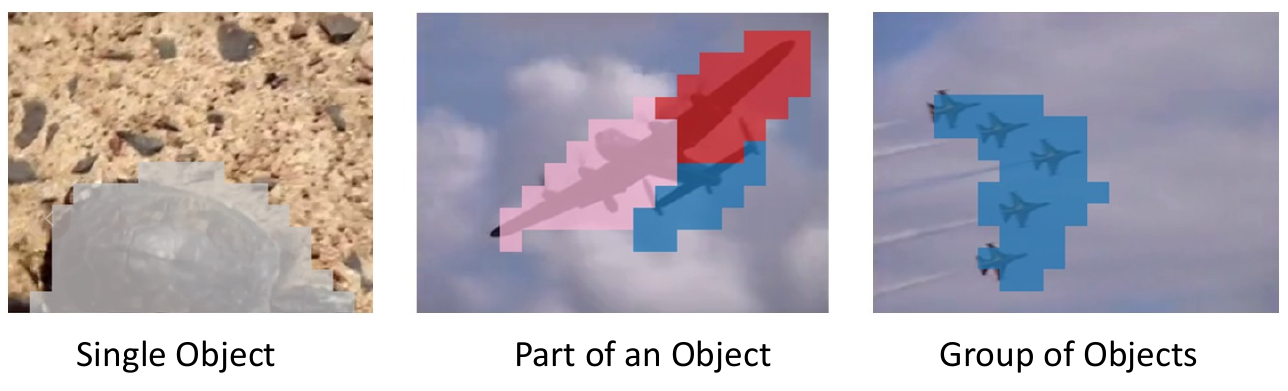}
    \caption{\textbf{Three types of slots.} We show examples of a slot containing a single object, part of an object or a group of objects. For simplicity we only colorize slots overlapping with objects, and slots in one image are colored differently.}
    \label{fig:slot_object_category}
\end{figure}
\section{Datasets, implementations and baselines}

\noindent\textbf{Implementation details.}
(1) \textit{Vision backbone.} We use the original VideoMAE module with two modifications: 1) we set each token to represent a $p\times p\times 1$ data patch instead of $p\times p\times 2$, and 2) we use 3D trigonometric function combination as the positional encoding. This modified VideoMAE is pre-trained on Something-Something-v2~\cite{goyal2017something} without supervision from human annotation.
(2) \textit{Patch-based CLIP} We train the patch-based CLIP model on ImageNet-1K~\cite{ILSVRC15} dataset without any class label. Specifically, we train a multi-head self-attention network to replace the last layer of the CLIP vision Transformer encoder. 

\noindent\textbf{Other baselines.}
For video slot learning, we compare against a number of OCL pipelines that output slots for each frame.
Those baselines do not perform slot association across frames and do not produce video slots (i.e., $K$ coherent tubes for a video).
Nevertheless, their output can be sent into the patch-based CLIP to be named. 
(1) \textit{Slot-Attention}~\cite{Locatello2020SlotAttention} uses a convolutional neural network to encode an image and extracts $K$ object-centric features using slot attention.
(2) \textit{SAVi}~\cite{kipf2022conditional}
is an extension of Slot Attention to videos that initializes the slots with some form of weak supervision, and then uses the slot from the previous frame to initialize the current frame.
Since our method is completely unsupervised, for a fair comparison, we do not provide this weak annotation in our evaluation.
(3) \textit{DINOSAUR}~\cite{dinosaur}
is a pixel grouping method that manages to group pixels on real-world data by freezing a pre-trained network as the backbone and reconstructing the extracted features.
Out of the three baselines, only DINOSAUR uses a pre-trained backbone (self-supervised with DINO~\cite{Caron2021DINO} on unlabeled ImageNet-1K).

\noindent\textbf{Datasets.}
(1) \textit{ImageNet-VID} (I-VID) is a common benchmark for video object detection which includes 30 object categories of the ImageNet DET dataset.
(2) \textit{YouTube-VIS} (Y-VIS) is a large-scale video instance segmentation dataset, which also includes object bounding boxes. 

\begin{table*}[ht] \small 
    \centering
    \adjustbox{max width=0.55\linewidth}{
    \begin{tabular}{c c c c c c c}
    \toprule
        \multirow{2}*{Method} & \multicolumn{3}{c}{ImageNet-VID} & \multicolumn{3}{c}{Youtube-VIS} \\
        \cmidrule(r){2-4} \cmidrule(r){5-7}
         & CorLoc & DecRate & mAP$_{50}$ & CorLoc & DecRate & mAP$_{50}$ \\ 
         \cmidrule(r){1-1} \cmidrule(r){2-4} \cmidrule(r){5-7}
        Slot-Attention  & 11.64 & 7.83 & 0.51$^*$ & 10.27 & 6.57 & 0.51$^*$ \\ 
        SAVi & 16.23 & 10.55 & 1.23$^*$ & 11.10 & 7.18 & 0.90$^*$ \\ 
        DINOSAUR &  48.04 & 39.58 & 7.34$^*$ & 44.80 & 34.63 & 8.15$^*$ \\ 
        \cmidrule(r){1-1} \cmidrule(r){2-4} \cmidrule(r){5-7}
        \textbf{Ours} &  \textbf{60.90} & \textbf{53.75} & \textbf{29.23} & \textbf{63.74} & \textbf{55.05} & \textbf{35.19} \\ \bottomrule
    \end{tabular}
    }
    \caption{\textbf{Comparison with other object-centric methods on object localization and slot labeling.} Note that numbers with $^*$ superscript means that the method uses our slot labeling module when computing the data. \label{tab:baselines}}
\end{table*}

\noindent\textbf{Metrics.}
To evaluate the localization performance of our method, we apply standard metrics in \textit{object localization} tasks.
Similar to previous studies~\cite{dinosaur,Vo2020discovery}, we report \textit{CorLoc}, which is the proportion of images where at least one object was correctly localized, and \textit{DecRate}, which measures the percentage of objects that are correctly detected out of all the ground-truth objects. 
To further evaluate the overall labeling performance, we use the traditional object detection metric mean Average Precision (mAP), which examines the accuracy of both object localization and semantic classification.

Besides, to understand how well our framework learns video slots, we quantify whether a slot indeed captures a single object.
Empirically, we observe that there are several other typical cases that deviate from that ideal behavior.
To quantitatively analyze the challenge posed by part-whole hierarchies, 
we categorize slots that overlap with objects based on whether they contain a single object (SO), part of an object (PO) or a group of objects (GO).
We label all those that do not overlap with any object or have only minor overlap with an object as background (BG).
See Figure~\ref{fig:slot_object_category} for a qualitative explanation and 
Appendix for details.
Please note that \emph{the classification of these categories is only meant to give a qualitative and informative assessment of the distribution of errors}.

\section{Experiments}

The complete design space is large, it includes, for video slot learning, 1) image (e.g., DINO) versus video backbone (e.g., VideoMAE) and 2) spatio-temporal versus per-frame grouping, and for video slot labeling, 3) vanilla CLIP versus patch-aligned adaptation and 4) joint optimization using both text and image features. This paper does not attempt to exhaust this space.
However, we will provide evidence that combining a video backbone with spatio-temporal grouping gives the best video slot learning ability, that CLIP adaptation is crucial, and that joint learning proves to be very effective. This conclusion is not at all unexpected. We now proceed to examine our (incomplete) evaluations with a set of questions.  

\noindent\textbf{Q1: How does our complete pipeline compare to the baselines?}
We present a quantitative evaluation on two datasets in Table~\ref{tab:baselines}.
As pixel-reconstruction methods, slot-attention and SAVi perform poorly on both real-world video datasets across all metrics, confirming the ineffectiveness of the pixel-space reconstruction loss.
DINOSAUR, which relies on feature reconstruction, performs better, particularly in terms of CorLoc and DecRate. However, its mAP score is only around 8, because it fails to align localization with semantics. In contrast, our model outperforms all other approaches across all metrics on both datasets, thanks to its ability to achieve spatio-temporal consistency and joint optimization with semantics.

\noindent\textbf{Q2: Does the choice of vision backbone and grouping algorithm matter for slot learning?}
\looseness-1As described in Section~\ref{sec:oclocalization}, video slot learning depends on the choice of vision backbone and grouping algorithm. Table~\ref{tab:ablation_stgrouping} demonstrates that a video-based backbone, such as VideoMAE, leads to superior performance in spatio-temporal slot learning across all metrics. Conversely, the spatial-only slot learning produces better results with an image-based backbone. Therefore, the combination of video backbone and grouping algorithm is crucial and must be selected accordingly.

\begin{figure*}[ht]
    \centering
    \includegraphics[width=0.85\linewidth]{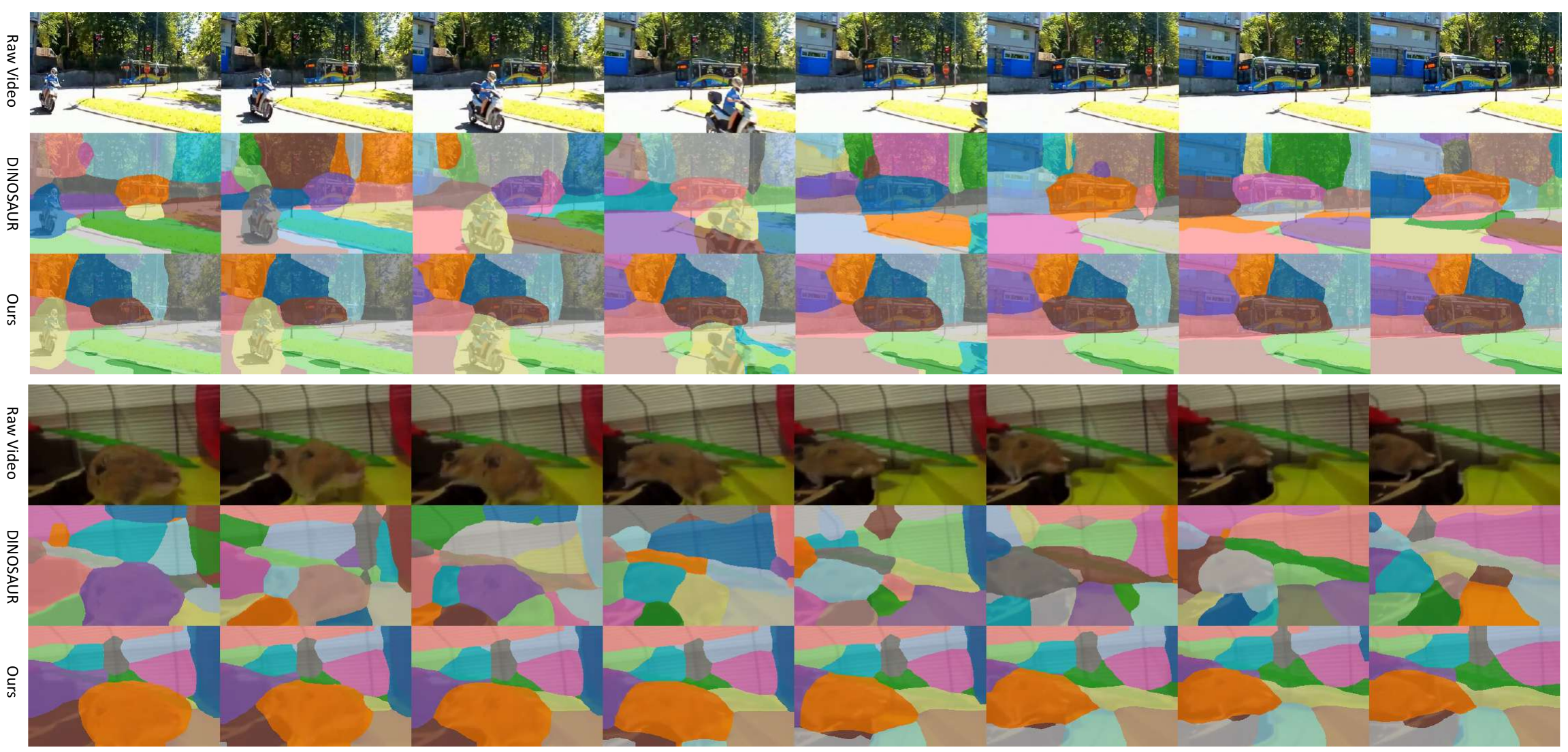}
    \caption{\textbf{Consistent localization in video.} We compare the localization performance from DINOSAUR and ours on two 8-frame video samples. We visualize the spatially interpolated localization for patch-based slot assignment. DINOSAUR is an image-based baseline thus its localization has no temporal consistency. In contrast, ours, as a spatio-temporal localization method, provides temporal consistent localization, as the same region or object are consistently localized by the same slot (indicated by the same color)  across frames.}
    \label{fig:video_slot_qualitative}
\end{figure*}

Although spatial DINO outperforms our model in terms of CorLoc and DecRate, we contend that these metrics pertain to image-based approaches. The higher mAP$_{50}$ score suggests that the temporal consistency ultimately improves the correct labeling and overall detection performance. To further illustrate the benefits of temporal consistency, we provide additional qualitative visualizations in Figure~\ref{fig:video_slot_qualitative}.
\begin{table} \small
    \centering
        \adjustbox{max width=0.95\linewidth}{
    \begin{tabular}{c c c c c}
    \toprule
        Backbone & Video Slot & CorLoc & DecRate & mAP$_{50}$ \\
        \cmidrule(r){1-1} \cmidrule(r){2-2} \cmidrule(r){3-5}
        \multirow{2}{*}{DINO} & $\times$   & 62.12 & 54.79 & 28.75 \\ 
        & $\checkmark$ & 59.84 & 52.98 & 28.40 \\ 
        \cmidrule(r){1-1} \cmidrule(r){2-2} \cmidrule(r){3-5}
        \multirow{2}{*}{VideoMAE} & $\times$  &  57.16 & 50.25 & 25.77 \\ 
        & $\checkmark$ & 60.90 & 53.75 & 29.23 \\
    \bottomrule
    \end{tabular}
    }
    \caption{\textbf{Ablations on vision backbones and video slots learning algorithm} on ImageNet-VID.}\label{tab:ablation_stgrouping}
    \vspace{-1em}
\end{table}

\noindent\textbf{Q3: How important is CLIP-based joint optimization?}
To address the importance of image-text alignment, we conduct a quantitative comparison of joint optimization with and without image-text alignment, as well as different design choices for alignment, using object localization and detection metrics. 
Our findings, summarized in Table~\ref{tab:ablation_pacl}, reveal that image-text alignment plays a critical role in joint optimization, with the choice of alignment method impacting all evaluation metrics. 
Notably, when joint optimization is applied, incorrect labeling using vanilla CLIP significantly impacts CorLoc and DecRate, while alignment using patch-based CLIP leads to significant improvements across all three metrics:
we can see from the first two rows only mAP is different, and the one with patch-based CLIP is better.
These results stress the importance of carefully tuning image-text alignment for optimal performance.
\begin{table} \small
    \centering
    \adjustbox{max width=0.95\linewidth}{
    \begin{tabular}{c c c c c c}
    \toprule
        Dataset & Joint & CLIP  & CorLoc & DecRate & mAP$_{50}$ \\ 
        \cmidrule(r){1-1} \cmidrule(r){2-2} \cmidrule(r){3-3} \cmidrule(r){4-6}
        \multirow{4}{*}{I-VID} & \multirow{2}{*}{\scalebox{0.75}{$\times$}} & Vanilla & 42.99 & 34.40 & 0.43 \\ 
        &  & Patch-based & 42.99 & 34.40 & 5.20 \\ 
        \cmidrule(r){2-2} \cmidrule(r){3-3} \cmidrule(r){4-6}
        & \multirow{2}{*}{ $\checkmark$} & Vanilla & 15.65 & 13.73 & 3.04 \\ 
        &  & Patch-based & 60.90 & 53.75 & 29.23 \\ 
        \cmidrule(r){1-1} \cmidrule(r){2-2} \cmidrule(r){3-3} \cmidrule(r){4-6}
        \multirow{4}{*}{Y-VIS} & \multirow{2}{*}{\scalebox{0.75}{$\times$}} &  Vanilla & 39.36 & 30.31 & 0.77 \\ 
        & & Patch-based& 39.36 & 30.31 & 5.96 \\ 
        \cmidrule(r){2-2} \cmidrule(r){3-3} \cmidrule(r){4-6}
        & \multirow{2}{*}{$\checkmark$} & Vanilla & 25.54 & 21.27 & 11.08 \\ 
        &  & Patch-based & 63.74 & 55.05 & 35.19 \\ \bottomrule
    \end{tabular}}
    \caption{\textbf{Ablations on patch-based CLIP.} We use the VideoMAE backbone and spatio-temporal video slot to conduct this ablation.}\label{tab:ablation_pacl}
    \vspace{-1em}
\end{table}
\subsection{Part-whole problem with joint optimization}
To estimate the effectiveness of slot localization in identifying objects, we assess SO, PO, GO, and BG on both I-VID and Y-VIS datasets, and present the results in Table~\ref{tab:part-whole-metric}.
Since slots are a direct output of video slot learning, they are more inclined to focus on parts of objects, accounting for $24.05\%$ and $34.80\%$ of the results (the PO column), rather than the entire object, which only represents $8.87\%$ and $6.04\%$. We observe that slots rarely contain several objects (GO column), and instead, often capture background classes ($65\%$ and $58\%$; BG column).

The reason for the relatively low proportion of SO can be attributed to the fact that visual similarity and common-fate inductive bias can only offer limited information when describing an object as a whole.  Since semantics is not inherently present in the visual modality, it is challenging to capture the full meaning of an object with visual cues alone. 

Our joint optimization approach demonstrates a significant increase in the ratio of SO slots. As illustrated in Table~\ref{tab:part-whole-metric}, the percentage of SO slots rises to $49.33\%$ and $40.75\%$, respectively, after the optimization with slot labeling. The reduction in PO and BG ratio indicates the source of improvement. First, slots that were previously tracking a significant part of the same object (PO) are merged to form a larger new slot. Further, each BG slot either has no overlap to any object, or only has minor overlap to one object. The reduction in BG ratio indicates that some of the latter case are correctly labeled and eventually merged.

This serves as an evidence that the joint optimization directly alleviate the part-whole issue. It should be noted that the proportion of GO also slightly increases, indicating that the model occasionally merge slots too aggressively for multiple instances in the same semantic category. This phenomenon again reflects the challenging associated with part-whole hierarchies.

Regarding the accuracy of labeling, Table~\ref{tab:ablation_pacl} presents the mAP score, which takes into account both localization and semantic labeling performance. With the aid of named localization and joint optimization, the model achieves a substantial increase in mAP score of around $24$ ($5.20\xrightarrow{}29.23$) and $30$ ($5.96\xrightarrow{}35.19$), respectively. Although these results fall short of the SOTA performance achieved by models using labels \cite{chen2020memory,he2021end}, considering that our framework is entirely unsupervised, the results are encouraging.

\begin{table} \small
    \centering
    \begin{tabular}{c c c c c c}
    \toprule
        Dataset & Joint & SO & PO & GO & BG \\
        \cmidrule(r){1-1} \cmidrule(r){2-2} \cmidrule(r){3-5} \cmidrule(r){6-6}
        \multirow{2}{*}{I-VID}  & \scalebox{0.75}{$\times$}& 8.87\% & 24.05\% & 1.96\% & 65.13\% \\ 
         & $\checkmark$& 49.33\% & 20.42\% & 7.01\% & 23.24\% \\
        \cmidrule(r){1-1} \cmidrule(r){2-2} \cmidrule(r){3-5} \cmidrule(r){6-6}
        \multirow{2}{*}{Y-VIS}  & \scalebox{0.75}{$\times$}& 6.04\% & 34.80\% & 0.77\% & 58.39\% \\ 
         & $\checkmark$& 40.75\% & 28.32\% & 5.64\% & 25.29\% \\
    \bottomrule
    \end{tabular}
    \caption{\textbf{Statistics of video slots.} We counted the percentage for each type of slots. \textbf{SO}: Single Object. \textbf{PO}: Part of an Object. \textbf{GO}: Group of Objects. \textbf{BG}: Slots for not \textit{\textbf{annotated}} regions in the dataset.}\label{tab:part-whole-metric}
    \vspace{-1em}
\end{table}

\subsection{Qualitative case studies}

\noindent\textbf{Spatio-temporal video slots.}
Our model offers a significant advantage over ordinary image slots in the form of spatio-temporal video slots.  These slots consistently focus on the same part across frames, which ensures temporal consistency and improves object-centric localization in videos. Figure~\ref{fig:video_slot_qualitative} provides visualizations that highlight the differences between image-based and video-based object-centric localization on two videos. While DINOSAUR can localize objects at a certain level in each frame, its results do not account for temporal consistency. This is evident as one object may be localized by a different number of slots among frames, and the background area may contain a totally different slot segmentation. In contrast, our model produces consistent localization across frames for all slots, including the background ones. This demonstrates that video slots tend to track all content in a short video clip. Moreover, this tracking behavior also handles objects that move in and out of frames. In the first sample, the biker disappears after frame 5, and its yellow slot never appears again. Similarly, the house in the background gradually appears and occupies the purple slot from frame 3.

\begin{figure}[ht]
    \centering
    \includegraphics[width=0.9\linewidth]{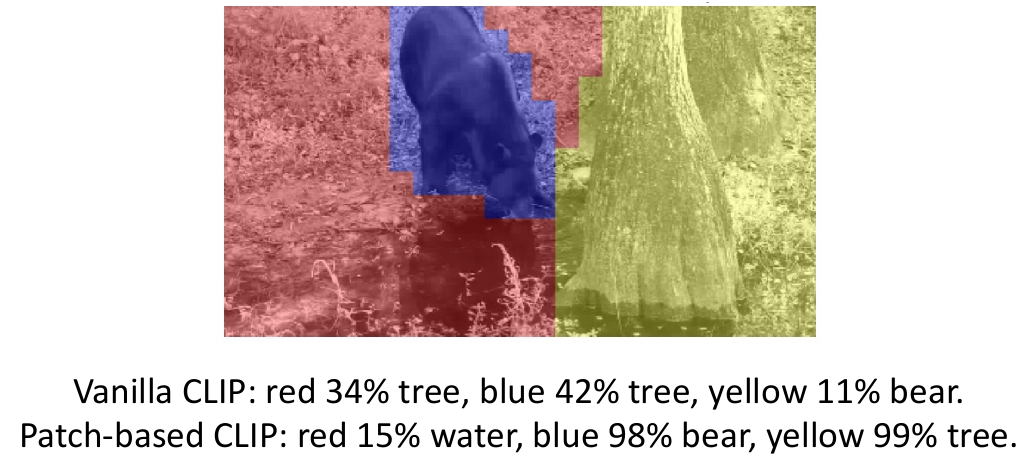}
    \caption{\textbf{Comparing patch-based CLIP and vanilla CLIP}. Vanilla CLIP cannot find the right name for the visual content in each slot, while patch-based CLIP names each slot correctly.}
    \label{fig:clip_fail}
\end{figure}

\noindent\textbf{Patch-based CLIP.}
Two qualitative examples comparing the labeling results from patch-based CLIP and vanilla CLIP are shown in Figure~\ref{fig:clip_fail}. The results clearly demonstrate that patch features from vanilla CLIP are inadequate for direct labeling algorithm, whereas the patch-based CLIP provides much better alignment and improved performance.

\begin{figure}[t]
    \centering
    \includegraphics[width=0.99\linewidth]{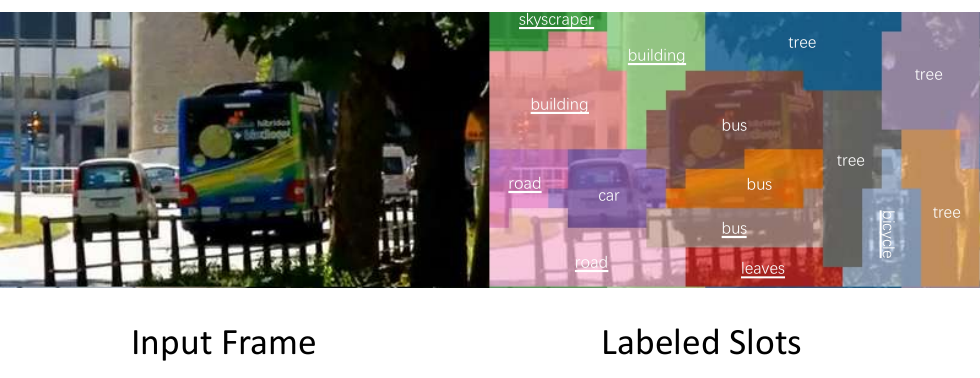}
    \caption{\textbf{Labeling any slot}. The first image shows the input frame. The second image shows the localization of that frame, as well as names for all slots. Slots names with underscore are those having a low cosine similarity.}
    \label{fig:name_any_slot}
    \vspace{-1em}
\end{figure}

\noindent\textbf{Labeling any slot.}
Given an appropriate list of semantics, our approach has the potential to name any slot. Figure~\ref{fig:name_any_slot} shows a single-frame example of the localization from video slots and their corresponding names. The result contains not only correct localization and labeling for the central objects, but also with reasonable names for other slots. Inevitably, there are slots that are not easy to name as their semantic features are not similar to any text features. Reasons of low similarity include 1) the precise name of the visual content is not included in the semantic list, 2) the visual content is a composition of multiple semantics (e.g. the red slot is a combination of leaves, road and fence), and 3) the visual content is incomplete or ambiguous (e.g. the light green slot only shows part of a building). Nevertheless, this example demonstrates both the potential of our model for open-set detection and segmentation, and at the same time also the challenging nature of these downstream tasks under an unsupervised setting.

\section{Discussion}

\noindent\textbf{On Supervision Signal.}
Our video object localization is entirely unsupervised apart from the implicit annotation contained in CLIP. Besides self-supervised pretraining and video slot learning, the patch-based CLIP finetuning is also without human annotation. Further, it is worth noting that either the pretrained or finetuned CLIP model do not generate any pseudo annotation to guide the learning process of the framework. The patch-based CLIP model is only used for label assignment at inference stage.

\noindent\textbf{Future work.}
Handling long videos is undeniably an intriguing task, but its comprehensive treatment lies beyond the scope of current work. However, we foresee potential extensions of our method from an engineering perspective in future research. For instance, the model can be readily expanded from processing a fixed number of frames to accommodating a larger and more flexible range of frames with relatively minor adjustments, such as position embedding interpolation or temporal frame sub-sampling. 
\section{Conclusion}

In this paper, we proposed a novel unsupervised method for object localization in videos that leverages the power of slot attention combined with a pre-trained CLIP image-text model.
By fine-tuning the last layer of CLIP in a self-supervised manner on unlabeled image data, we established a means for applying the image-text mapping to local patch tokens.
Through a joint optimization process with both visual and semantic information, we achieved, for the first time, high-quality and spatio-temporally consistent object localization outcomes on real-world video datasets.
While the current coarse localization provides an encouraging starting point, we acknowledge that it still suffers from low resolution of the patch tokens.
Moreover, to complete our object detection and tracking pipeline, we are yet to move beyond semantic localization and accurately differentiate between individual object instances.
Nevertheless, we are optimistic that our findings will pave the way for label-free training on complex video analysis tasks

{\small
\bibliographystyle{ieee_fullname}
\bibliography{camera_ready}
}
\clearpage
\appendix

\begin{appendices}
\section{Datasets and metrics}
\subsection{Datasets}
We utilize Something-Something-v2~\cite{goyal2017something} to pre-train our vision backbone and use ImageNet-1K ~\cite{Deng2009ImageNet} to fine-tune the patch-based CLIP model. To train and evaluate video slots learning and labeling, we opt for two datasets, ImageNet-VID~\cite{russakovsky2015imagenet} and YouTube-VIS~\cite{yang2019video}, which focus on video objects.

% and ImageNet-1K ~\cite{Deng2009ImageNet} to pre-train our vision backbone and patch-based CLIP model, respectively. 

\paragraph{Something-Something-v2} is a vast video collection showcasing individuals performing actions in specific environments. Something-Something-v2 contains around 169k training videos and 25k validation videos from 174 classes. Each video features both an action and object, and merely considering appearance information would not suffice. Thus, we employ the video files to pre-train our VideoMAE, disregarding any annotations.

\paragraph{ImageNet-VID} is a commonly used benchmark for video object detection. The dataset comprises 3862 and 555 video snippets in its training and validation sets, respectively. The video streams are annotated on every frame at either 25 or 30 fps. Furthermore, it includes 30 object categories, which are a subset of the categories in the ImageNet DET dataset.

\paragraph{YouTube-VIS} is a large-scale video instance segmentation dataset, including 2,883 high-resolution YouTube videos with pixel-level annotations of object instances. The videos are diverse in terms of scene complexity, motion, and object categories, making it a challenging dataset for video object tasks. Although initially designed for instance segmentation, YouTube-VIS includes object bounding boxes, making it compatible with video object detection. Specifically, since the annotation of YouTube-VIS validation set and test set is inaccessible, we make a customized train-val split randomly.

\subsection{Metrics}
\paragraph{Detailed definition of type of slots (SO/PO/GO/BG)}
The part-whole hierarchies has been a long-existing issue in object-centric learning~\cite{hinton2022represent}. As for a visually complex object, the localization has the chance to split it in multiple slots. We categorize slots that overlap with objects as containing a Single Object (SO), Part of an Object (PO) or a Group of Objects (GO). We utilize intersection, area, and union of the slots and objects to determine their relations and help categorizing slots.

Formally, for each predicted bounding box $b_{p}$ derived from one slot and all the ground truth bounding box $b_{g}$, we have the following rules to categorize this slot with respect to two threshold parameters $\tau_{1}$ and $\tau_{2}$:

\begin{enumerate}
    \item If the intersection over union (IoU) between the two boxes are larger than $\tau_{1}$, we regard it as Single Object. If there exists only one $b_{g}$ such that the intersection over $b_{g}$'s area is larger than $\tau_{2}$, we also regard it as Single Object.
    \item Else, if the intersection over $b_{p}$'s area is larger than $\tau_{2}$, we regard it as Part of an Object.
    \item Else, if there exists multiple $b_{g}$ such that the intersection over each $b_{g}$'s area is larger than $\tau_{2}$, we regard it as Group of Objects.
    \item Finally, if none of the above conditions hold, the slot should be Back-Ground slot. 
\end{enumerate}

% This case would occur when a large slot covers a small object so that the IoU could be small.
In practice, we set $\tau_{1}=0.5$ and $\tau_{2}=0.5$.
Based on the definition, we propose to use the proportion of each type of slots as a metric to evaluate how well the object-centric model can handle the part-whole problem.

\section{Implementation details}
\subsection{Training of vision backbone}
We conduct the experiments with 128 NVIDIA Tesla T4 GPUs for pre-training of VideoMAE on Something-Something V2 datasets. Specifically, we use ViT-Base as our encoder and a four-layer transformer as our decoder. The patch size is set to $16 \times 16 \times 1$. The videos are sampled to length of 16 and cropped to $224\times 224$. The videos are reconstructed given the $90\%$ patches masked. 

For optimizer, we utilize AdamW~\cite{loshchilov2017decoupled} with a base learning rate of $1.5e-4$. We pre-train the model for 800 epochs with 40 warm-up epochs and cosine learning rate scheduler. The total batch size is 512.
\subsection{Training of video slot learning}
We train the video slot learning module using the Adam optimizer \cite{kingma2014adam} with a learning rate of $3e^{-3}$, linear learning rate warm-up of 5 000 optimization steps and an exponentially decaying learning rate schedule. Further, we clip the gradient norm at 1 in order to stabilize training. The experiments on ImageNet-VID and YouTube-VIS are conducted separately. We train the model on specific dataset for 300 epochs. The models were trained on 8 NVIDIA Tesla T4 GPUs with a local batch size of 10, where each sample contains 8 frames. The number of slots is set to 15.

\paragraph{Training with variable resolution} There are some special designs to handle variable resolution in these datasets. For the data, we first resize the short side of the video into 224 while keeping the aspect ratio. Then we resize its long side to the nearest multiple of patch size to make it fit the patchify operation in the ViT model. For different models involved in the pipeline, we employ positional embedding interpolation techniques in both vision backbone (VideoMAE) and patch-based CLIP. As for the video slot grouping module, slot attention is naturally feasible to handle variable length tokens. We make similar positional embedding interpolation in the slot decoder part. During training, we random sample frames repeatedly from one video for each batch to avoid the issue of batching variable length tokens.

\subsection{The finetuning of patch-based CLIP}
For patch-based CLIP, we finetune a pre-trained ViT-B/16 CLIP model on ImageNet 1K dataset, which includes 1.28M images. For each image, we augment the training with a random resized crop with scale between 0.9 and 1 and resize the cropped area to 224$\times$ 224, then a random horizontal flip. The model is finetuned for 200 epochs on 32 NVIDIA Tesla T4 GPUs with a local batch size of 4096. The optimizer is SGD, and the learning rate scheduler is cosine annealing with initial learning rate of 1.

\subsection{Inference pipeline}
In this part, we give a detailed explanation of the full inference pipeline. Given a video $V\in\mathbb{R}^{T\times H \times W\times 3}$ as input, we explicitly divide it into three steps. 

\textit{(1) Video Slot Extraction.} We first patchify the video and extract its spatiotemporal patch features with shape $\mathbb{R}^{T \times H' \times W' \times D}$ by the vision backbone, where $T$ is number of frames, $H$ and $W$ are the height and width of the video, $H'$ and $W'$ are the height and width after patchify operation, $D$ is the hidden dimension. Next, we feed the patch features to the trained spatiotemporal grouping module to get the patch-slot assignment $\alpha \in \mathbb{R}^{K\times T \times H' \times W'}$, where $K$ is the number of slots. The patch-slot assignment serves as the preliminary localization.

\textit{(2) Slot Labeling.} We feed the same video $V$ to the patch-based CLIP frame by frame to get the semantic patch features with shape $\mathbb{R}^{T \times H' \times W' \times D}$. Next, the patch-slot assignment $\alpha$ is employed to aggregate the semantic patch features into semantic slots features with shape $\mathbb{R}^{K\times D}$ by average pooling. Then we use the semantic slots features to match a set of text features with shape $\mathbb{R}^{Q\times D}$ extracted from text prompts, where $Q$ is the number of text prompts. Next, we take the top probability as the semantic label of the slot and get the named localization.

\textit{(3) Joint Optimization.} Given the named localization from last step, we further process them with the joint optimization. As described in the main paper, the joint optimization is mainly responsible for excluding some slots of not interests and merge some neighbor slots with the same semantic label.

\section{Qualitative results}

\subsection{Failure cases analysis}

We visualize and discuss four typical failures or mistakes by our model.

\begin{figure}[h]
    \centering
    \includegraphics[width = \linewidth]{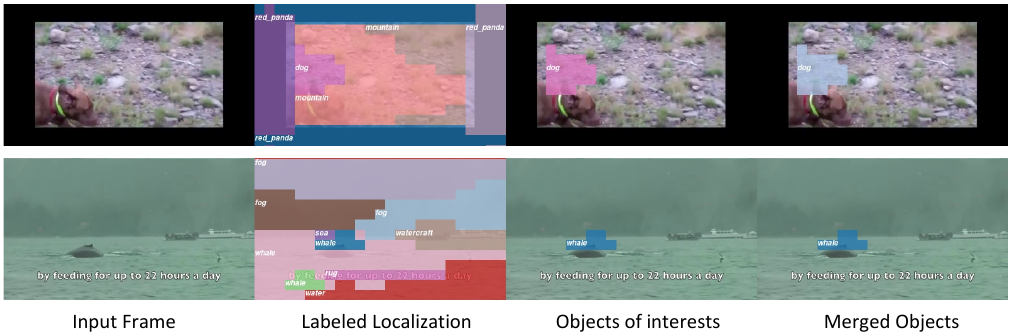}
    \caption{Examples of localization failure.\label{fig_localization}}
\end{figure}

\paragraph{Localization failure}
This type of failure is mainly caused by inaccurate segmentation of our models. In each samples in Fig.~\ref{fig_localization}, since some characteristic features of the object are included in one slot that overlaps partially with the object, \textit{e.g.} the ear of the dog or the dorsal fin of the whale, our model still recognizes their name, and that leads to inaccurate localization but correct naming.
%our model cannot precisely localize the dog and whale. However, since some characteristic features of the object (the ear of the dog, or the dorsal fin of the whale) are included, our model can still recognize their name, which leads to inaccurate localization but correct naming.

\begin{figure}[h]
    \centering
    \includegraphics[width = \linewidth]{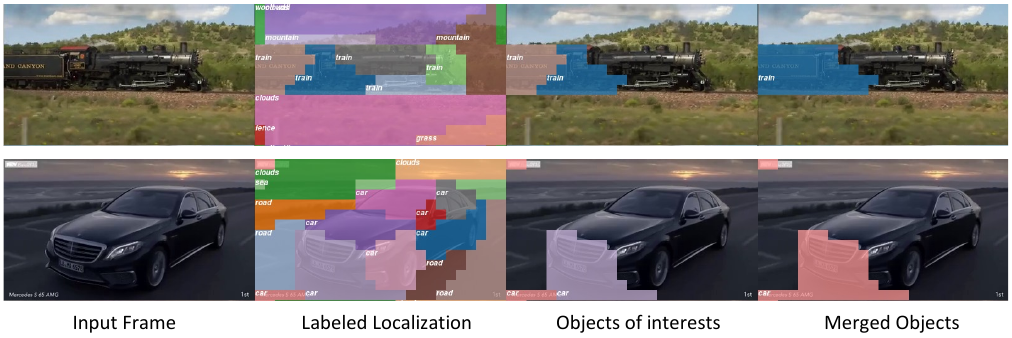}
    \caption{Examples of background removal mistakes.\label{fig_foreground}}
\end{figure}

\paragraph{Background removal mistakes}
Our model removes two types of background slots in post-processing: the slots named with common background labels and the slots resembling neither target nor common background labels. 
However, the pipeline may wrongly remove the object of interest. For example, in Fig.~\ref{fig_foreground}, the car and the train are over-segmented into several parts. Although our patch-based CLIP can recognize the class of the patch, however, the cosine similarity of such slots with the text vector may be small. In such circumstances, those parts of the target object may be removed, leading to incomplete objects after merging. 

\begin{figure}[h]
    \centering
    \includegraphics[width = \linewidth]{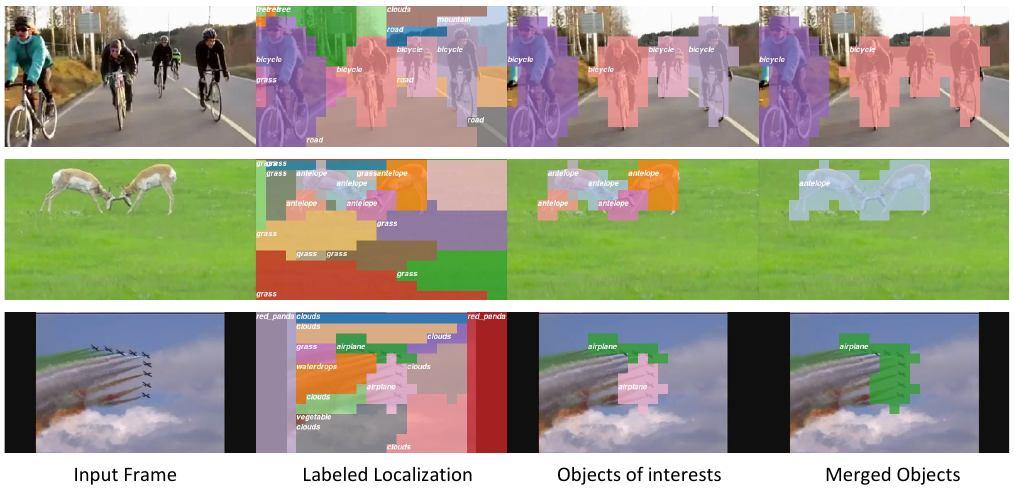}
    \caption{Examples of multiple instance failure.\label{fig_multi_instance}}
\end{figure}

\paragraph{Multiple instance failure}
Limited by the insensitivity of CLIP model to quantity words, we cannot distinguish how many objects (especially objects of the same type) in a slot. The merging process tends to merge objects with the same label into groups of objects. For example, in Fig.~\ref{fig_multi_instance}, bicycles and antelopes are merged together respectively in each sample due to the shared semantic meaning and close spatial relationship across multiple instances.

Moreover, due to the resolution limit of slot attention mechanism, our model cannot always segment adjacent small objects.  For example, the small airplanes are grouped together by the slot attention in the last row of Fig.~\ref{fig_multi_instance}.

\begin{figure}[h]
\centering
\includegraphics[width = \linewidth]{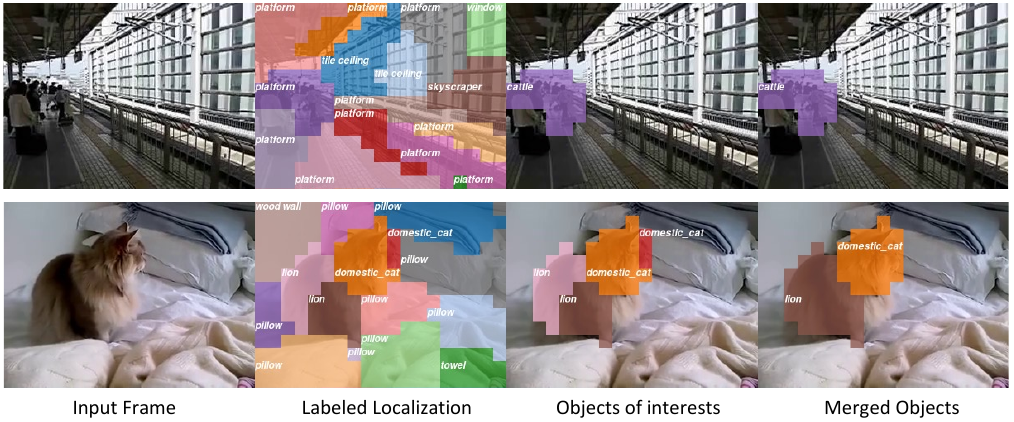}
\caption{Examples of naming mistakes.\label{fig_name}}
\end{figure}

\paragraph{Naming mistake}
The patch-based CLIP may not be able to recognize every slot. On one hand, the word may have multiple meanings, in the first row Fig.~\ref{fig_name}, the group of people is named as ``cattle'', due to cattle also means ``human beings especially en masse''. On the other hand, the patch-based CLIP may not be able to recognize the slot if the slot is just a part of an object. In the second row Fig.~\ref{fig_name}, the hair of cat is recognized as lion, which is also reasonable since the hair may also come from a lion. 

\subsection{More qualitative examples}
\begin{figure*}[htb]
    \centering

    \adjustbox{width=\linewidth,trim=6em 0 0em 0}{
    \includegraphics{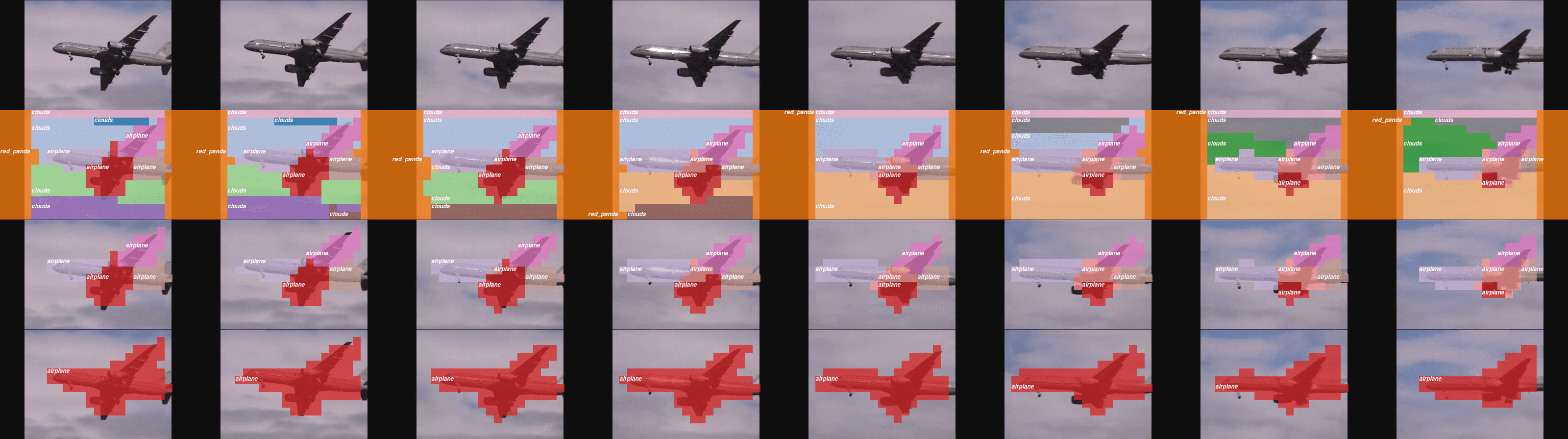}
    }

    \adjustbox{width=\linewidth,trim=6em 0 0em 0}{
    \includegraphics{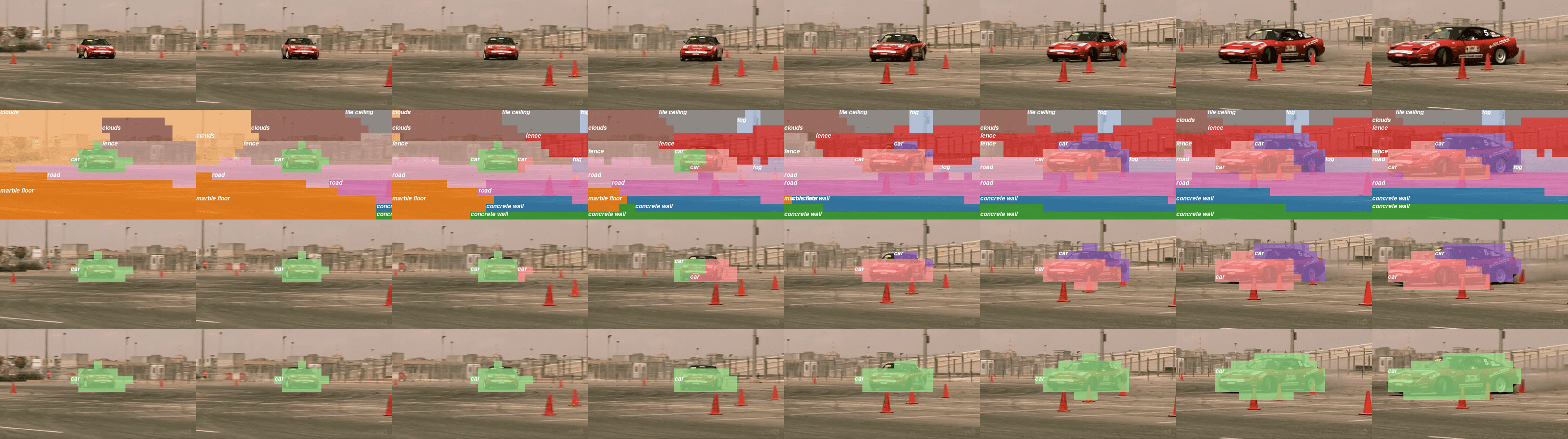}
    }

    \adjustbox{width=\linewidth,trim=6em 0 0em 0}{
    \includegraphics{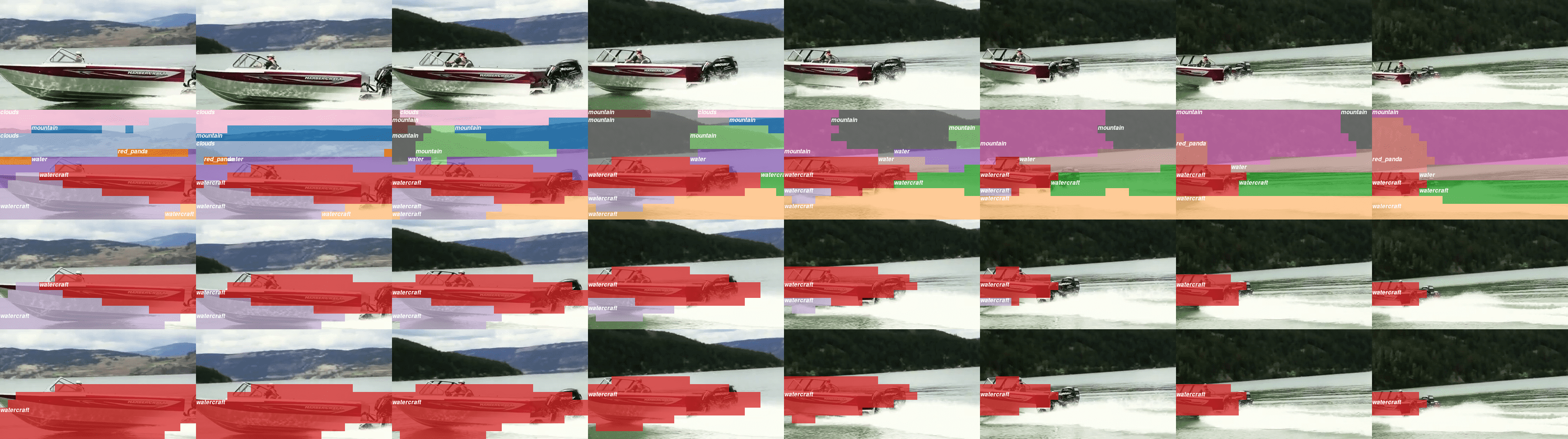}
    }
    \adjustbox{width=\linewidth,trim=6em 0 0em 0}{
    \includegraphics{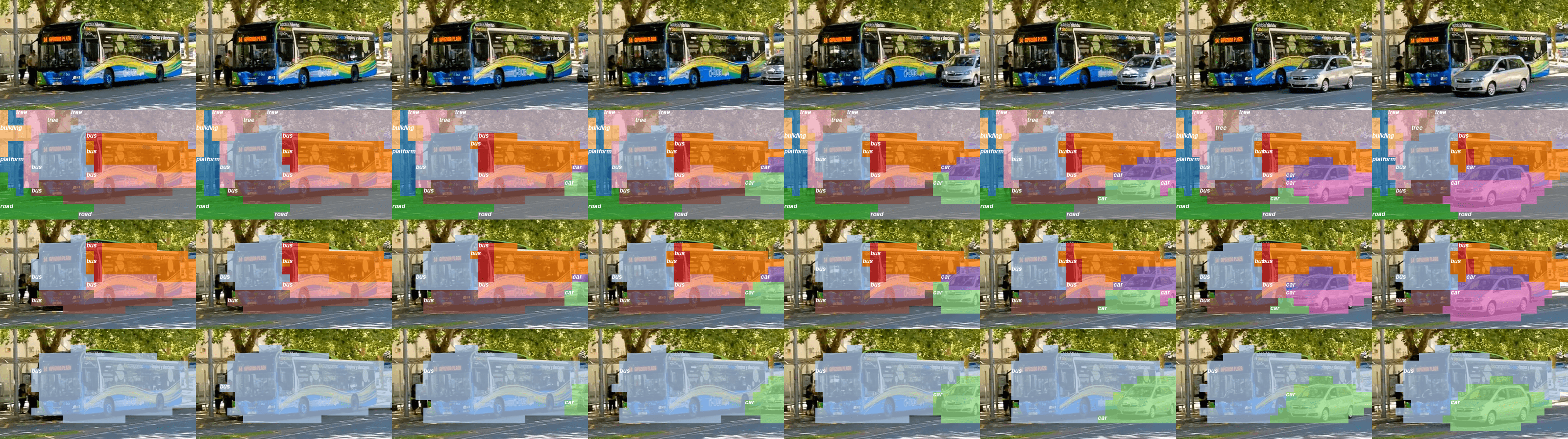}
    }
    \caption{More qualitative examples on ImageNet-VID dataset. The four rows in each example correspond to the input frames, labeled localization, objects of interests and merged objects.\label{fig_imagenet_first}}
\end{figure*}

\begin{figure*}[htb]
    \centering
    \adjustbox{width=\linewidth,trim=6em 0 0em 0}{
    \includegraphics{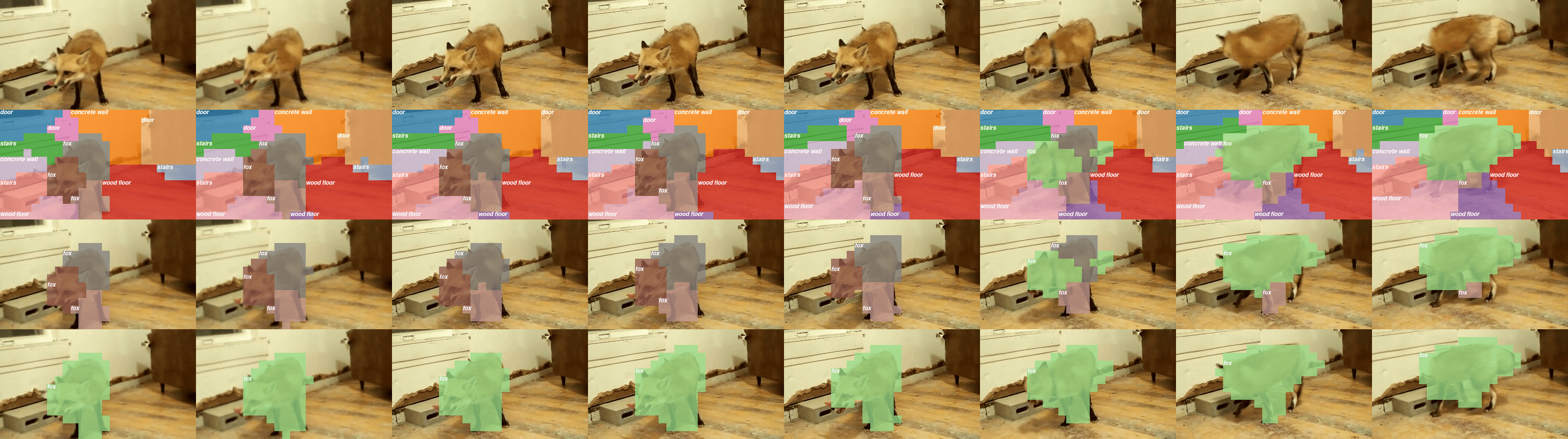}
    }
    \adjustbox{width=\linewidth,trim=6em 0 0em 0}{
    \includegraphics{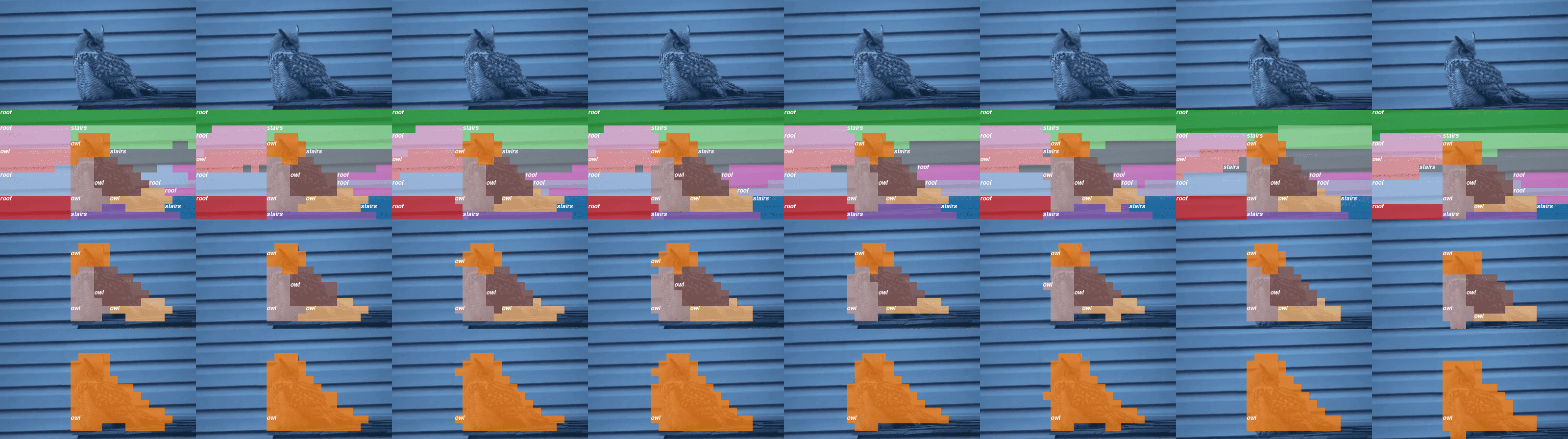}}
    
    % \adjustbox{width=\linewidth,trim=6em 0 0em 0}{
    % \includegraphics{iccv2023AuthorKit/figs/supp_ytvis/val_video_7_num_4_horizontal_text.png}}
    
    \adjustbox{width=\linewidth,trim=6em 0 0em 0}{
    \includegraphics{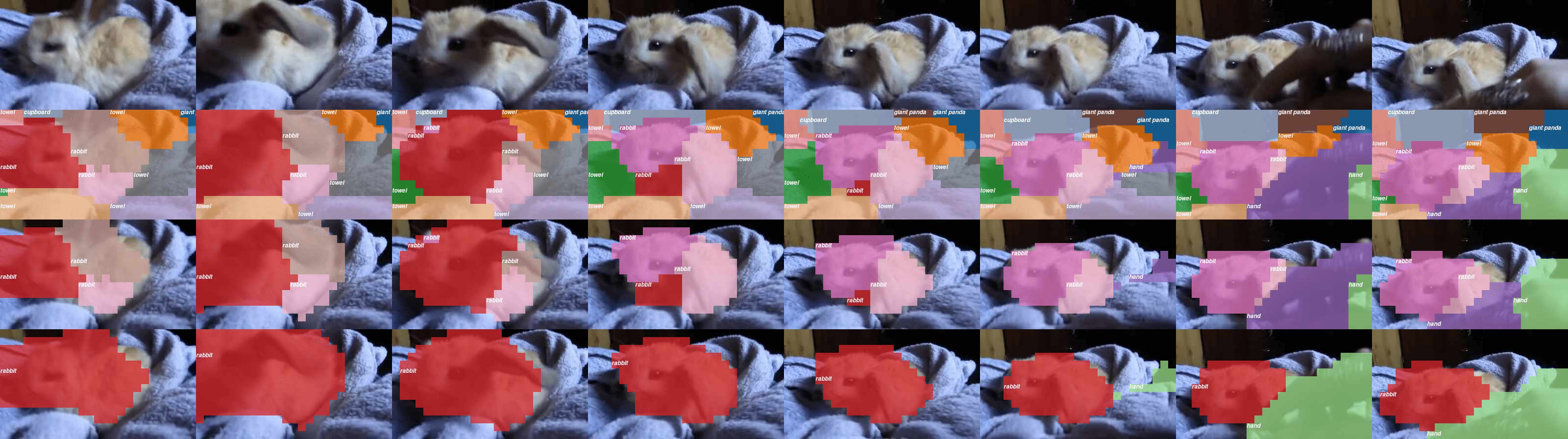}}

    \adjustbox{width=\linewidth,trim=6em 0 0em 0}{
    \includegraphics{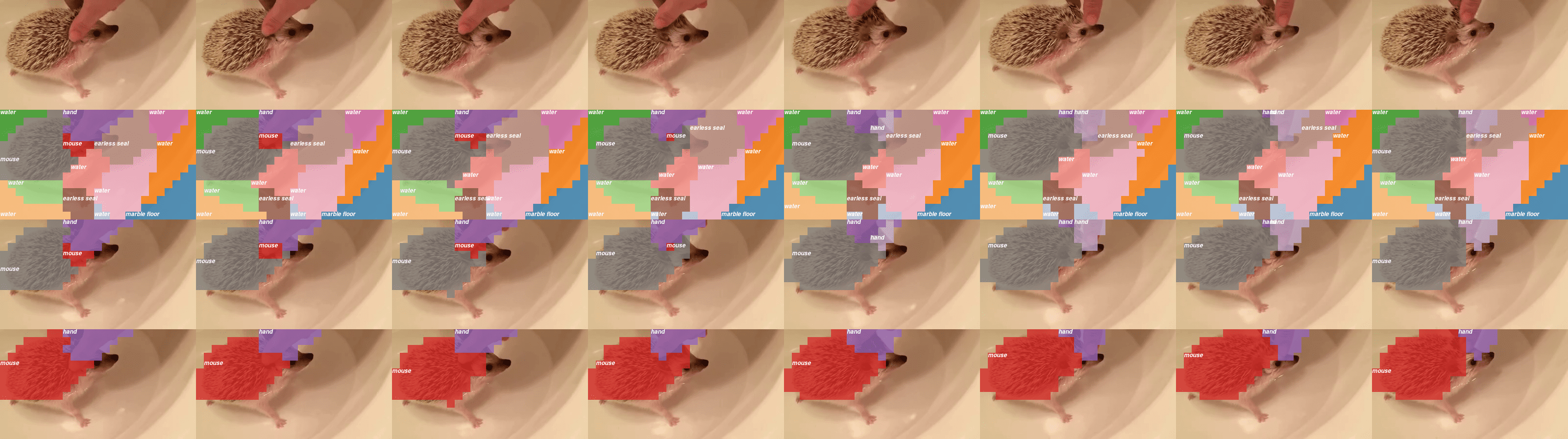}}

    \caption{
    More qualitative examples on YouTube-VIS dataset. The four rows in each example correspond to the input frames, labeled localization, objects of interests and merged objects.\label{fig_ytvis}}
\end{figure*}

We show more qualitative examples on ImageNet-VID and YouTube-VIS datasets in Figure~\ref{fig_imagenet_first} and Figure~\ref{fig_ytvis}, respectively.

\section{Prompt engineering}
To fully utilize the language models, we do not directly use the embedding of the words for classes in Youtube-VIS or ImageNet-VID as the text features. But instead, we embed each class name into the sentence with its form as ``a photo of a [CLASS]''. For \textit{common background labels}, we select the stuff classes of COCO-stuff~\cite{caesar2018coco}. Generally, stuff classes correspond to amorphous background regions such as grass and sky. 

Some minor modifications are made to transfer the format of some classes names, e.g. from ``wall-concrete'' to ``concrete wall''. Interestingly, though we add the article ``a'' in the text template, our model can still match to patches with multiple objects. To some extent, this can further help explain why our model doesn't distinguish groups of same-category instances together. We also tried adding some synonyms of some classes, for example,  ``ape'' for ``monkey'', and ``impala'' for ``antelope''. However, the improvement of synonyms is not significant and we report the results without synonyms.
\clearpage

\end{appendices}

\end{document}